\pdfoutput=1

\documentclass[11pt]{article}

\usepackage{acl}
\usepackage{CJKutf8}

\usepackage{times}
\usepackage{latexsym}

\usepackage[T1]{fontenc}

\usepackage[utf8]{inputenc}

\usepackage{microtype}
\usepackage{booktabs}
\usepackage{tabularx}
\usepackage{multirow}
\usepackage{multicol}

\usepackage{float}
\usepackage{amsmath}
\usepackage{amsfonts}
\usepackage{verbatim}
\usepackage{graphicx}
\usepackage{relsize}
\usepackage{url}
\usepackage{xspace}
\usepackage{cleveref}
\usepackage{adjustbox}
\usepackage{arydshln}
\usepackage{array}
\usepackage{xcolor}
\usepackage{tcolorbox}
\definecolor{mycolor}{HTML}{285291}
\definecolor{highlight}{HTML}{E9BCEB}
\definecolor{darkertext}{HTML}{285291}
\definecolor{background}{HTML}{FAFAFA}
\newcommand{\sd}[1]{\smaller[2]\ensuremath{\pm#1}}
\newcommand{\tabref}[1]{Table~\ref{#1}\xspace}
\newcommand{\figref}[1]{Figure~\ref{#1}\xspace}

\newcommand{\appref}[1]{Appendix\xspace\ref{#1}\xspace}
\newcommand{\myparagraph}[1]{\noindent\textbf{#1}\xspace}

\newcommand{\task}[1]{\textbf{#1}\xspace}
\newcommand{\taskt}[1]{\textit{#1}\xspace}

\title{Can Large Language Model Comprehend Ancient Chinese? \\A Preliminary Test on ACLUE}

\author{Yixuan Zhang \quad Haonan Li\\
Mohamed bin Zayed University of Artificial Intelligence, UAE \\
haonan.li@mbzuai.ac.ae
}

\begin{document}
\maketitle
\begin{CJK*}{UTF8}{gbsn}
\begin{abstract}

Large language models (LLMs) have showcased remarkable capabilities in understanding and generating language. However, their ability in comprehending ancient languages, particularly ancient Chinese, remains largely unexplored. To bridge this gap, we present ACLUE, an evaluation benchmark designed to assess the capability of language models in comprehending ancient Chinese. ACLUE consists of 15 tasks cover a range of skills, spanning phonetic, lexical, syntactic, semantic, inference and knowledge.
Through the evaluation of eight state-of-the-art LLMs, we observed a noticeable disparity in their performance between modern Chinese and ancient Chinese. Among the assessed models, ChatGLM2 demonstrates the most remarkable performance, achieving an average score of 37.4\%. We have made our code and data public available.\footnote{\url{https://github.com/isen-zhang/ACLUE}}

\end{abstract}

\section{Introduction}
The study of ancient languages provides valuable insights into the past civilizations' thoughts, languages, societies, and histories \cite{zhiming1990language,woodard2008ancient,DBLP:journals/pnas/Bouchard-Cote0G13}. Ancient China, as one of the oldest civilizations, has left a significant impact on contemporary societies including Japan, Korea, and Vietnam. However, existing research in ancient Chinese language processing have primarily focused on specific time periods or genres \cite{DBLP:conf/acl/YanLHZ16,DBLP:conf/acl-alta/XieLC19,DBLP:journals/talip/LiuYQL20,hu2021knowledge,AnchiBERT}. Typically, the previously proposed models require customized fine-tuning for particular tasks.

Recently, the significant advancements made in large language models (LLMs) underscore their remarkable proficiency across a range of tasks, showcasing their potential in performing various tasks without the need for fine-tuning \cite{gpt3,bloom,touvron2023llama,bloomz,glm-130b}. These models encapsulate extensive knowledge and sophisticated reasoning capabilities. Notably, the emergence of ChatGPT~\cite{gpt-4} and Chinese-oriented LLMs such as ChatGLM~\cite{glm-130b}, has accentuated their remarkable ability in comprehending and generating modern language. However, due to the lack of ancient language benchmarks, the abilities of LLMs in handling ancient language remains largely unexplored.

We present the Ancient Chinese Language Understanding Evaluation (ACLUE), an evaluation benchmark consisting of 15 tasks. These tasks are derived from a combination of manually curated questions from publicly available resources, and automatically generated questions from classical Chinese language corpora. The range of questions span from the Xia dynasty (2070 BCE) to the Ming dynasty (1368 CE), covering a broad temporal range. Similar to the well-established LLM benchmarks such as ARC \cite{arc} and MMLU \cite{mmlu}, ACLUE adopts multiple-choice question format for all tasks.  This ensures simplicity and uniformity in evaluating models, accommodating variations in different training or fine-tuning procedures and prompting methodologies.

In our preliminary experiments, we assessed the performance of 8 advanced LLMs, where the Chinese LLM ChatGLM2 demonstrates the best performance with an average accuracy of 37.4\%, slightly surpassing ChatGPT. However, considering the baseline accuracy of 25\% from random guessing and the average accuracy of around 50\% achieved by the same models on contemporary modern Chinese benchmarks such as AGIEval \cite{AGIEval} and CMMLU \cite{li2023cmmlu}, we believe there is still ample room for improvement in the proficiency of existing LLMs in understanding ancient Chinese.

\section{ACLUE Benchmark}

ACLUE consists of 15 tasks that encompassing lexical, syntactic, semantic, inference, and general knowledge of ancient Chinese. The details of the tasks are provided in \appref{app:tasks}, where basic statistics can be found in \tabref{tab:data_subjects}, and examples of each task are listed in \tabref{tab:data_example}.
The questions cover a wide range of genres, including poetry, prose, classical novels, couplets, historical records, and biographies, spanning the period from 2070 BCE to 1368 CE. Among the 15 tasks, 8 were automatically generated using existing corpora or datasets, 5 were collected from freely available standard tests, and 2 were directly sourced from other work. Each task includes 100 to 500 questions, exceeding the number required for testing a human participant. 

ACLUE serves as an evaluation suite for LLMs ability in understanding ancient Chinese without task-specific fine-tuning.
To ensure fair comparison among different models trained with varying approaches, all tasks are formatted into multiple-choice questions with four choices, of which only one is correct. The task details and dataset construction process are elaborated in this section. 

\subsection{Lexical Tasks}

We create three lexical tasks using the ancient Chinese corpus, which includes over 50,000 word sense annotations and 3,000 named entity annotations \citep{shu-etal-2021-gu}.

\myparagraph{Polysemy resolution} aims to understand the different senses or meanings of words. Two types of questions are created: one asks which character in a given sentence carries a particular meaning, while the other requires identifying the meaning of a character within the sentence.

\myparagraph{Homographic character resolution} focuses on recognizing homographic characters in ancient Chinese texts. Homographic characters, also known as ``通假字'' (tōng jiǎ zì) in Chinese, are substitutions of characters in ancient Chinese texts with others that have similar pronunciation or appearance. 

\myparagraph{Named entity recognition} focuses on identifying named entities (e.g., names of people, places, dynasties, etc.) in ancient Chinese texts. Two types of questions are created: one type asks for the specific entity type of a given entity within a contextual sentence, while the other type asks in which context a Chinese word represents an entity.

\subsection{Syntactic and Semantic Tasks}

\myparagraph{Sentence segmentation} is a task that involves choosing the correct segmentation of a given sentence. 
Since ancient Chinese lacks punctuation marks, accurate sentence segmentation becomes crucial for analyzing syntax and semantics of a sentence.
We create the task by sampling sentences from the Classical-Modern Chinese Corpus,\footnote{\url{https://github.com/NiuTrans/Classical-Modern}} which provides labeled sentence segmentation. To create false options, we manipulate the original punctuation marks by moving, adding, or deleting them.

\myparagraph{Couplet prediction} involves predicting the most likely second line of a Chinese couplet based on a given first line. Chinese couplet, also known as ``对联'' (duì lián), is a traditional form of poetic expression consisting of two lines of verse. The two lines are expected to match in terms of meaning, rhyme, and other poetic elements. We construct this task using a couplet dataset.\footnote{\url{https://github.com/wb14123/couplet-dataset}}

\myparagraph{Poetry context prediction} is a task constructed using the Chinese-poetry corpus.\footnote{\url{https://github.com/chinese-poetry}} The objective of this task is to select the most likely next or previous sentence given a specific sentence from a poem. 

\subsection{Inference}

\myparagraph{Poem quality estimation} task is constructed based on dataset proposed by \citet{Yimrl:18}, which consists of 173 Chinese quatrains, with each one being rated for fluency, coherence, and meaningfulness on a scale of 0 to 5 by human expert.
We randomly select four poems and create questions asking models to identify the best or worst poem based on a specific criterion. To ensure clear distinctions, we maintain a minimum score differences of 2 between the correct option and the other options. The task aims to evaluate the ability of models to compare the quality of Chinese quatrains.

\myparagraph{Reading comprehension} is based on the AGIEval dataset \cite{AGIEval}. It contains a subset of Chinese Gaokao questions. We select questions that contains ancient Chinese text from this subset.

\myparagraph{Poetry sentiment analysis} involves predicting the sentiment of an entire poem or parts of a poem, determining whether it is positive, neutral, or negative. We utilize a dataset proposed by \citet{DBLP:conf/cikm/ShaoSWWG21}, which contains 5,000 poems. Each poem and its individual sentences are labeled with fine-grained sentiment categories, including negative, implicit negative, neutral, implicit positive, and positive sentiments. We merge implicit negative and implicit positive labels with their respective categories to address ambiguity.

\myparagraph{Poetry appreciation} is manually curated from openly accessible online resources.

\subsection{Knowledge-intensive Tasks}

Ancient Chinese knowledge tasks cover various subjects, including \task{ancient Chinese medical}, \task{ancient Chinese literature}, \task{traditional Chinese culture}, and \task{ancient Chinese phonetics}. To create these tasks, we collected relevant questions from various online open resources.
Additionally, we extracted a subset of questions from the CMMLU dataset \cite{li2023cmmlu}, which consist of questions at the high-school level in current Chinese education. This selection allows us to form the tasks of \task{basic ancient Chinese}.

\begin{table*}[t]
\small
    \centering
    \begin{adjustbox}{width=1\textwidth}
    \begin{tabular}{@{}lccccccccccccccccccccc@{}l}
    \toprule
    \multirow{2}{*}{Model} & \multicolumn{3}{c}{Lexical} && Syntactic && \multicolumn{2}{c}{Semantic} &&\multicolumn{4}{c}{Inference}&& \multicolumn{5}{c}{Knowledge}&& \multicolumn{2}{c}{\multirow{2}{*}{Overall}} \\
\cmidrule{2-4}\cmidrule{6-6}\cmidrule{8-9}\cmidrule{11-14}\cmidrule{16-20}
 &T1&T2&T3&&T4&&T5&T6&&T7&T8&T9&T10&&T11&T12&T13&T14&T15&&\\\midrule
ChatGLM2 & \textbf{45.4} & 24.4 & 34.8 & & \textbf{46.4} & & 39.8 & 24.6 & & 28.3 & 29.7 & \textbf{42.7} & \textbf{52.6} & & 28.9 & \textbf{50.7} & 34.6 & 43.8 & \textbf{35.0} && \textbf{37.4} & \sd{8.9}\\
ChatGPT & 41.8 & 20.6 & \textbf{41.2} & & 43.0 & & 45.4 & 27.4 & & \textbf{39.7} & \textbf{39.6} & 38.8 & 47.8 & & 29.3 & 43.4 & 34.6 & 33.8 & 27.0 && 36.9 & \sd{7.6}\\
BLOOMZ & 45.2 & 22.4 & 35.6 & & 32.2 & & \textbf{60.2} & 27.2 & & 31.5 & 17.8 & 26.2 & 45.2 & & 29.7 & 44.1 & \textbf{39.3} & \textbf{44.4} & 29.0 && 35.3 & \sd{10.7}\\
ChatGLM & 39.6 & 19.4 & 39.4 & & 36.6 & & 37.2 & 23.4 & & 30.8 & 32.7 & 30.1 & 43.8 & & 29.3 & 36.8 & 30.8 & 40.6 & 27.0 && 33.2 & \sd{6.6}\\
Falcon & 40.4 & \textbf{28.8} & 21.2 & & 32.6 & & 37.2 & \textbf{31.4} & & 36.9 & 22.8 & 31.1 & 43.8 & & \textbf{30.5} & 30.1 & 30.3 & 36.9 & 26.0 && 32.0 & \sd{6.0}\\
Baichuan & 31.6 & 26.4 & 22.0 & & 33.0 & & 37.2 & 27.8 & & 30.3 & 16.8 & 25.2 & 38.2 & & 27.3 & 36.0 & 37.0 & 41.9 & 31.0 && 30.8 & \sd{6.5}\\
LLaMA & 36.4 & 22.2 & 26.4 & & 33.0 & & 29.6 & 29.6 & & 31.5 & 18.8 & 24.3 & 41.8 & & 24.5 & 23.5 & 29.4 & 29.4 & 31.0 && 28.8 & \sd{5.6}\\
MOSS & 30.6 & 27.6 & 25.8 & & 24.0 & & 30.0 & 25.0 & & 29.8 & 27.7 & 21.4 & 30.8 & & 26.5 & 22.1 & 24.6 & 22.5 & 26.0 && 26.3 & \sd{3.0}\\
\bottomrule
    \end{tabular}
    \end{adjustbox}
    \caption{Zero-shot average accuracy of all models. The overall results are averaged (with standard deviation) over all tasks. T1: Polysemy resolution, T2: Homographic character resolution, T3: Named entity recognition, T4: Sentence segmentation, T5: Couplet prediction, T6: Poetry context prediction, T7: Poetry quality estimation, T8: Reading comprehension, T9: Poetry appreciation, T10: Poetry sentiment analysis, T11: Basic ancient Chinese, T12: Traditional Chinese culture, T13: Ancient Chinese medical, T14: Ancient Chinese literature, T15: Ancient Chinese phonetics.}
    \label{tab:main_results}
\end{table*}
\vspace{-0.2cm}

\begin{figure}[t]
	\scriptsize
	\begin{tcolorbox}[colframe=white, left=3mm, right=3mm]
   以下是关于\colorbox{highlight}{古代文学知识}的单项选择题，请直接给出正确答案的选项。\\
    \textcolor{mycolor}{Here are some multiple-choice questions about \colorbox{highlight}{Ancient Chinese} \colorbox{highlight}{literature}, please provide the correct answer choice directly.}\\\\
    题目：下列诗句中，属于杜牧咏史诗的是：\\
    \textcolor{mycolor}{Question: Among the following lines of poetry, the one that belongs to Du Mu's historical poem is:} \\
    A. 旧时王谢堂前燕，飞入寻常百姓家\xspace\\ \textcolor{mycolor}{In former times, the swallows in front of the halls of Wang and Xie flew into the homes of ordinary people}\\
    B. 长空澹澹孤岛没，万古销沉向此中\xspace\\ \textcolor{mycolor}{The vast sky engulfed the desolate island, and for eternity it sank into this place.}\\
    C. 千寻铁锁沉江底，一片降幡出石头\xspace\\ \textcolor{mycolor}{Thousands of chains sank to the bottom of the river, and a stone emerged with a descending flag}\\
    D. 三百年间同晓梦，钟山何处有龙盘\xspace\\ \textcolor{mycolor}{For three hundred years, the same dream awakened at dawn, where on Zhongshan Mountain can a dragon coil}\\
    答案是：\xspace \textcolor{mycolor}{(Answer:)}
	\end{tcolorbox}
	\caption{An examples from ACLUE. English translations are provided for better readability.}
	\label{fig:example}
\end{figure}

\section{Experiment}
To provide an overview of the language ability of existing open-sourced LLMs on ancient Chinese, we assess 8 models including 4 multilingual models: ChatGPT~\cite{gpt-4}, LLaMA~\cite{touvron2023llama}, Falcon~\cite{falcon40b}, BLOOMZ~\cite{bloomz}, and 4 Chinese models: ChatGLM~\cite{du2022glm}, Baichuan,\footnote{https://github.com/baichuan-inc/baichuan-7B} ChatGLM2~\cite{glm-130b}, and MOSS~\cite{moss}.
Details about these models are introduced in \appref{app:models}.

For models optimized to function as chatbots, such as ChatGPT and ChatGLM, we generate output and use regular expressions to extract the answer key. 
For other models, we directly obtain the probability of the next tokens after the prompt and selected the one with the highest probability among the answer keys (i.e., `A',`B',`C',`D').
We employ both zero-shot (do not provide examples) and in-context five-shot (provide few examples) evaluation.
An example of evaluation instance is shown in \figref{fig:example}.

\subsection{Results}
\tabref{tab:main_results} shows the zero-shot performance of all models. The five-shot results are similar to the zero-shot results, suggesting that models can comprehend the task without additional demonstrations. 
Overall, the Chinese model ChatGLM2 demonstrates the best performance, with an average accuracy of 37.4\%. Moreover, its performance on almost all tasks is above the random guessing (25\%).
The multilingual model ChatGPT achieves a slightly lower accuracy of 36.9\%, compared to ChatGLM2, yet it maintains relatively consistent performance in terms of standard deviation.

Regarding specific tasks, we have several findings: (1) BLOOMZ exhibits exceptional performance in \taskt{couplet prediction} (T5), achieving an accuracy of 60.2\%. This accuracy is nearly double that of most other models, possibly due to BLOOMZ's training set, xP3, having overlaps with our data source.
Similar, ChatGLM2 may have been exposed to the original texts used for \taskt{sentence segmentation} (T4) and \taskt{poetry appreciation} (T9), which explains its proficient performance in these tasks. 
(2) All models face challenges in the \taskt{homographic character resolution} (T2), with performance close to random guessing. This issue likely arises because the auto-regressive training objective does not emphasize understanding of homographic concepts.
(3) \taskt{Reading comprehension} (T8) poses a considerable challenge for all models due to the extreme long length of the question (nearly 1,000 tokens on average). Specifically, BLOOMZ, LLaMA, and Baichuan are significantly affected, exhibiting lower performance on this task compared to their average across other tasks. This observation suggests that these models may lack adequate support for processing very long input.

\begin{figure}[t]
    \centering
    \includegraphics[width=0.5\textwidth]{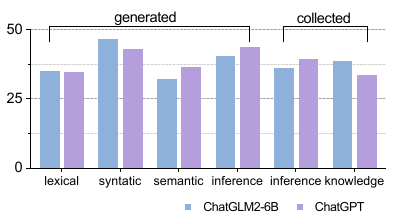}
    \caption{The performance of ChatGPT and ChatGLM2 on ACLUE of different categories.}
    \label{fig:compare}
\end{figure}

Based on data origin, we divide the tasks into two categories: auto-generated and manually collected. In \figref{fig:compare}, we compare the performance of ChatGPT and ChatGLM2, the best multilingual and Chinese models, respectively. We find that while ChatGLM2 exhibits superior overall performance on ACLUE, its dominance only observed in the auto-generated syntactic tasks and collected knowledge categories. More comparison results are provided in \appref{app:gen_col}.

In terms of data quality and reliability, auto-generated questions within ACLUE were slightly less intricate than collected questions, but the difference was not significant. This suggests that the auto-generated questions hold reasonable potential for effectively evaluating models' grasp of ancient Chinese language.

\section{Related Work}

A lot of research has been conducted on various aspects of ancient Chinese language processing, encompassing topics such as ancient Chinese to modern Chinese translation \cite{DBLP:journals/talip/LiuYQL20}, Chinese couplets generation \cite{DBLP:conf/acl/YanLHZ16,DBLP:conf/mipr/YuanZLZ19,DBLP:journals/soco/Qu0L022}, Classic Chinese poem generation \cite{DBLP:conf/cncl/YiLS17,DBLP:conf/emnlp/YangSYL18,DBLP:conf/acl/GuoYSLYLCZL19,DBLP:conf/acl-alta/XieLC19,DBLP:conf/dmbd/ZhaoBWW22,DBLP:conf/eacl/MaZW23}, and ancient Chinese sentence segmentation \cite{DBLP:journals/corr/abs-1810-03479,hu2021knowledge}, as well as general language model pre-training \cite{AnchiBERT}. 
However, many of these studies focus on specific types or literary formats that were popular during specific time periods. 

Recently, large language models have demonstrated remarkable language understanding and generation capabilities \cite{gpt3,bloom,falcon40b}. Researchers have began to evaluate these LLMs based on their performance across a wide range of tasks \cite{touvron2023llama,bloomz,gpt-4}.
However, the absence of a comprehensive evaluation benchmark poses a challenge in assessing the performance of LLMs in ancient language understanding. 
Existing ancient Chinese evaluation datasets either have a narrow focus on specific tasks, limiting the scope of evaluation, or require model fine-tuning prior to evaluation.
In contrast, ACLUE provides a natural support for evaluation under zero-shot and in-context learning settings, making it more compatible with LLMs.

\section{Conclusion}

We propose ACLUE, the first evaluation benchmark for ancient Chinese language understanding. Our preliminary evaluation of 8 large language models reveals that, despite their exceptional performance in modern language understanding, they struggle with even basic tasks in ancient Chinese. 
Through analysis, we illustrate that the auto-generated questions possess similar difficulty levels to those found in actual school tests. 

\bibliography{anthology,custom}
\bibliographystyle{acl_natbib}
\appendix
\section{Data details}\label{app:tasks}
The \tabref{tab:data_subjects} listed the Chinese, category, and origin of the tasks in ACLUE, and the \tabref{tab:data_example} provides examples for each task.

\begin{table*}[t]
\small
    \centering
    \begin{tabular}{lccccc}
    \toprule
    Task & Total Q. & Avg. len & Task (zh) & Category & Origin\\
    \midrule 
    Named entity recognition & 500 & 138& 古汉语命名体识别& lexical&generated\\
    Polysemy resolution & 500 & 116& 古文单字多义 &lexical&generated\\
    Homographic character resolution & 500 & 137& 通假字& lexical&generated\\
    Sentence segmentation & 500 & 210& 古文断句 &syntatic&generated\\
    Couplet prediction & 500 & 62& 对联预测 &semantic&generated\\
    Poetry context prediction & 500 & 77& 古诗词上下句预测 &semantic&generated\\ 
    Poetry sentiment analysis & 500 & 60 & 诗词情感分类 &inference&generated\\
    Poem quality estimation & 406 & 118 & 古诗词质量评估 &inference&generated\\
    Ancient Chinese medical & 211 & 38 & 医古文 &knowledge&collected\\
    Ancient Chinese literature & 160 & 44& 古代文学知识 &knowledge&collected \\
    Traditional Chinese culture & 136 & 59& 国学常识 &knowledge&collected\\
    Poetry appreciation & 103 & 258& 古诗词曲鉴赏 &inference&collected\\
    Basic ancient Chinese & 249 & 52 & 基础古汉语知识 &knowledge&collected\\
    Reading comprehension & 101 & 982& 古文阅读理解& inference&collected\\
    Ancient Chinese phonetics & 101 & 50 & 古音学 &knowledge&collected\\
    \bottomrule
    \end{tabular}
    \caption{ACLUE task overview. We list the total number of questions (Total Q.), average question length counted in Chinese characters (Avg. len), task names in Chinese, task type, and data origin type. }
    \label{tab:data_subjects}
\end{table*}
\section{Further analysis}\label{app:gen_col}
\begin{figure}[t]
    \centering
    \includegraphics[width=0.5\textwidth]{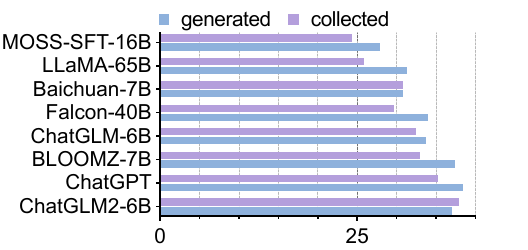}
    \caption{The performance comparison of LLMs on ACLUE across different data origins.}
    \label{fig:gen_col}
\end{figure}

The performance comparison of all LLMs on different data origins is illustrated in \figref{fig:gen_col}. Evaluating the LLMs' performance on auto-generated questions versus manually collected questions in ACLUE, we found that while the generated questions were less intricate than the collected ones, the difference was not significant. This indicates a comparable level of difficulty between the two types of questions. Among all the models, only ChatGLM2 demonstrated better performance on collected questions compared to auto-generated questions, which may indicate exposure to the original question texts used in ACLUE.

\section{Models being Evaluated}\label{app:models}
\paragraph{BLOOMZ} is derived from BLOOM through fine-tuning on a crosslingual task mixture (xP3), which is an instruction-following dataset. BLOOMZ exhibits competitive performance with models that have a larger number of parameters across various non-generation tasks.

\paragraph{Baichuan-7b} is an open-source large-scale pre-trained model developed by Baichuan Intelligence. Built on the Transformer architecture, it adopts the same model design as LLaMA. This 7-billion-parameter model was trained on approximately 1.2 trillion tokens using proprietary Chinese-English bilingual corpora, with optimization focused on Chinese. 

\paragraph{ChatGLM-6B} is bidirectional dense model pre-trained using the General Language Model (GLM) algorithm developed by Tsinghua University. It supports bilingual (Chinese and English) language processing. ChatGLM is a version of GLM that has been supplemented with supervised fine-tuning, feedback bootstrap, and reinforcement learning with human feedback, specifically optimized for Chinese question answering (QA) and dialogue tasks. 

\paragraph{ChatGLM2-6B} is the second generation of ChatGLM. It uses the hybrid objective function of GLM, and has undergone pre-training with 1.4T bilingual tokens and human preference alignment training. It offers enhanced performance and an expanded context length of 32K. With efficient inference using Multi-Query Attention technology, it achieves efficient inference with higher speed and lower memory usage.

\paragraph{ChatGPT} is a GPT model developed by OpenAI and fine-tuned using reinforcement learning from human feedback (RLHF). As a commercial product, specific details about its model size, training data, and training process are not disclosed.

\paragraph{LLaMA-65B} is an auto-regressive language model proposed by Meta. It incorporates several structural improvements over the vanilla transformer and is trained on a mixture of publicly available data sources. LLaMA has demonstrated comparable or even superior performance to models that are ten times its size.

\paragraph{Falcon-40B} is a decoder-only model created by TII and trained on 1,000B tokens of RefinedWeb \cite{refinedweb} data. Due to the high quality of its training data, Falcon-40B performs competitively with LLaMA-65B on various benchmarks.

\paragraph{MOSS} is an open-source Chinese language model proposed by Fudan University. It matches ChatGPT in terms of training scale and alignment techniques. MOSS-SFT is initialized with CodeGen and further pre-trained on 100B Chinese tokens and 20B English tokens. The SFT (supervised fine-tuned) version of MOSS-SFT enables the model to follow instructions in multi-turn dialogues.

\begin{table*}[t]
\scriptsize
    \centering
    \begin{tabular}{llp{10cm}}
    \toprule
    ID & Task & Example\\
    \midrule 
    T1 & 单字多义 & 下列选项中对“此神农之所以[长]，而尧舜之所以章也。”这句话中的“长”字理解正确的是( )\\
    & Polysemy resolution    & \textcolor{darkertext}{\textbf{A. 首领}} \hspace{0.5cm}B.排行第一, 长子\hspace{0.5cm}C.长处, 专长\hspace{0.5cm}D. 长大，成年\\[5pt]
    T2 & 通假字 & 列选项中[]内的“红”字是通假字的是( )\\
    & Homographic character resolution & \textcolor{darkertext}{\textbf{A. 吾已食禄，又夺园夫女[红]利虖。}}\hspace{0.5cm}B. 晓看[红]湿处，花重锦官城\\
    && C. [红] 芳满院参差折，绿醑盈杯次第衔。\hspace{0.5cm}D. 竹缘浦以被绿，石照涧而映 [红] 。\\[5pt]
    T3 & 命名体识别 & 下列选项中[]内的“阳”字代表了地名的是( )\\
    & Named entity recognition    &\textcolor{darkertext}{\textbf{A. 夫人授兆丹书真文、月中玉珰。}} \hspace{0.5cm}B. 令飞升上造洞[阳]之宫。\\
    &    &C. 今朝日[阳]里，梳落数茎丝。 \hspace{0.5cm}D. 晓发碧水[阳]，暝宿金山寺。\\[5pt]
    T4 & 古文断句 & 以下选项断句正确的是( )\\
    & Sentence segmentation    & \textcolor{darkertext}{\textbf{A. 史记/曰/秦使武安君白起攻赵/赵发兵拒秦/秦大破赵於长平/}} \\
    &    & B. 史记/曰秦/使武安君白起攻赵/赵发兵拒秦/秦大破赵於长平 \\
    &   &C. 史记/曰/秦使武安君白起攻赵/赵发兵拒秦秦大/破赵於长/平\\
    &    & D. 史/记曰秦使武安君白起攻赵/赵发兵拒秦秦大/破赵於长平\\[5pt]
    T5 & 对联 & “兔去龙来，交替人间春好景”的下联最可能是( )\\
    &Couplet prediction    &\textcolor{darkertext}{\textbf{A. 莺歌燕舞，和谐社会岁祥光。}}\hspace{0.5cm}B. 香遗书案，传家苦读育春风。\\
    &    &C. 赛龙夺锦，万人江岸闹端阳。\hspace{0.5cm}D. 情牵天下，凭谁设榻效陈蕃。\\[5pt]
    T6 & 古诗词上下句预测 & “何當秣馬候明發，便可一葦横長江。”的上一句是( )\\
    &Poetry context prediction    &\textcolor{darkertext}{\textbf{A. 江頭藉草作寒食，細雨梨花思故鄉。}}\hspace{0.5cm}B. 孰知文有忌，情至自生哀。\\
    &    & C. 樽前誰唱醉翁曲，鳥歌花舞催紅粧。\hspace{0.5cm}D. 千村萬落鳥呼客，山南嶺北花吹香。\\[5pt]
    T7 & 古诗词质量评估 & 下列古诗词前后文连贯性最差的是( )\\
    &Poem quality estimation    &\textcolor{darkertext}{\textbf{A. 阴雨难侵牖|春虫足哺儿|年年秋报喜|牛女有佳期}}\\
    &    &B.富贵良非愿|林泉毕此生|酒因随量饮|诗或偶然成\\
    &    &C.久不闻山歌|南风五月多|牧童呼伴侣|吹笛下西坡\\
    &    &D.今日骐驎阁|当年鹦鹉洲|寄书愁不达|书达得无愁\\[5pt]
    T8 & 古文阅读理解 & 下列对原文有关内容的理解和分析，表述不正确的一项是( )\\
    & Reading comprehension & 谢贞，字元正，陈郡阳夏人，晋太傅安九世孙也。父蔺，正员外郎，... ... 察因启曰：“贞有一子年六岁。”即有敕长给衣粮。（节选自《陈书·列传第二十六》，有删改）。【注】惠连：谢惠连，南朝宋文学家。\\
    &    &A. 谢贞天性聪慧，小时候读过不少典籍，有的读过就能背诵，有的粗通大意；他八岁时写的诗就深得长辈称赞。\\
    &    &B. 谢贞受府长史周确委托，为他撰写辞让都官尚书的表文。陈后主读过之后，怀疑该表文不是周确亲笔所作。\\
    &    &\textcolor{darkertext}{\textbf{C. 谢贞非常孝顺，小时候祖母因病难以进食，他便也不进食；父亲去世他悲痛欲绝，之后，奉养母亲未曾间断。}}\\
    &    &D. 母亲去世后，谢贞一心守丧，极度悲痛，骨瘦如柴，令人叹息。他忧病而死后，后主下令长期供他儿子吃穿。\\[5pt]
    T9 & 古诗词曲鉴赏 &下列对这首诗的赏析，不正确的一项是( )\\
    &Poetry appreciation &《幽居初夏》陆游。湖山胜处放翁家，槐柳阴中野径斜。水满有时观下鹭，草深无处不鸣蛙。箨龙已过头番笋，木笔犹开第一花。叹息老来交旧尽，睡来谁共午瓯茶。\\
    &    &A. 首句“湖山”二字总冒全篇，勾勒环境，笔力开张，巧妙地从山光水色中引出“幽居”。\\
    &    &\textcolor{darkertext}{\textbf{B. 首句概言“湖山胜处”，颔联写湖，是近处宽处静景；颈联写庭院周围，是远处细处动态。}}\\
    &    &C. 诗中写放翁心中郁结与柳宗元《小石潭记》中写“以其境过清”时的心境相似。\\
    &    &D. 本诗前三联写景，尾联结情，景情相衬，描写与抒情紧密关联，脉络清晰。\\[5pt]
    T10 & 诗词情感分类 & 古诗词“庭前芍药妖无格|池上芙蕖净少情|唯有牡丹真国色|花开时节动京城“的整体情感是( )\\
    & Poetry sentiment analysis & \textcolor{darkertext}{\textbf{A. 积极的}} \hspace{0.5cm}B. 消极的\hspace{0.5cm}C. 中性的\hspace{0.5cm}D. 无法判断\\[5pt]
    T11 & 国学常识 & "近朱者赤，近墨者黑"所蕴含的道理和下列哪句话最相似？( )\\
    & Basic ancient Chinese    &A. 青出于蓝，而胜于蓝。\hspace{0.5cm}\textcolor{darkertext}{\textbf{B. 蓬生麻中，不扶而直。}}\\
    && C. 公生明，偏生暗。\hspace{0.5cm}D. 三天打鱼两天晒网\\[5pt]
    T12 & 古汉语知识 & 下列句中，含有双宾语的一句是( )\\
    &Traditional Chinese culture    &A. 夫晉何厭之有? \hspace{0.5cm}\textcolor{darkertext}{\textbf{B. 重爲之禮而歸之。}}\hspace{0.5cm}C. 兔不可復得，而身爲宋國笑。\hspace{0.5cm}D. 甚矣，汝之不惠!\\[5pt]
    T13 & 医古文 & 以下除（）之外，都有病愈之义。 \\
    &Ancient Chinese medical    & A. 已\hspace{0.5cm} B. 起\hspace{0.5cm} \textcolor{darkertext}{\textbf{C. 性}}\hspace{0.5cm} D. 差\\[5pt]
    T14 & 古代文学知识 & 杜甫《春望》中的“感时花溅泪，恨别鸟惊心”所反映的是( )\\ 
    & Ancient Chinese literature    & A. 早年的读书和漫游生活。\hspace{0.5cm}B. 困居长安十年时的感受。 \\
    &    & \textcolor{darkertext}{\textbf{C. “安史之乱”时的国恨家愁。}}\hspace{0.5cm}D. 晚年漂泊西南的客旅生活。\\[5pt]
    T15 & 古音学 & 下列字在古代的声母、调类、等和开合口标注错误的是( )\\
    & Ancient Chinese phonetics    &\textcolor{darkertext}{\textbf{A. 温（影母平声二等开）}}\hspace{0.5cm}B. 权（群母平声三等合） \\
    &    &C. 空（溪母平声一等合）\hspace{0.5cm}D. 狂（群母平声三等合）\\[5pt]
    \bottomrule
    \end{tabular}
    \caption{ACLUE tasks examples.}
    \label{tab:data_example}
\end{table*}

\end{CJK*}
\end{document}